\documentclass[sigconf]{acmart}

\AtBeginDocument{%
  \providecommand\BibTeX{{%
    \normalfont B\kern-0.5em{\scshape i\kern-0.25em b}\kern-0.8em\TeX}}}

\acmConference[WiSeML '22]{WiSeML '22: 15th ACM Conference on Security and Privacy in Wireless and Mobile Networks}{May 16--19, 2022}{San Antonio, Texas, USA}
\acmDOI{10.1145/3522783.3529530}

\usepackage{enumitem}
\begin{document}
\fancyhead{}

\title{MR-iNet Gym: Framework for Edge Deployment of Deep Reinforcement Learning on Embedded Software Defined Radio}

\author{Jithin Jagannath\textsuperscript{1}, Kian Hamedani\textsuperscript{1}, Collin Farquhar\textsuperscript{1}, Keyvan Ramezanpour\textsuperscript{1}, Anu Jagannath\textsuperscript{1}}
\affiliation{%
  \institution{
  \textsuperscript{1}Marconi-Rosenblatt AI/ML Innovation Lab, ANDRO Computational Solutions LLC} 
  \country{Rome, NY, USA}
}
\email{{jjagannath, khamedani, cfarquhar, kramezanpour, ajagannath}@androcs.com}
\renewcommand{\shortauthors}{Jagannath and Hamedani, et al.}

\begin{abstract}
 Dynamic resource allocation plays a critical role in the next generation of intelligent wireless communication systems. Machine learning has been leveraged as a powerful tool to make strides in this domain. In most cases, the progress has been limited to simulations due to the challenging nature of hardware deployment of these solutions. In this paper, for the first time, we design and deploy deep reinforcement learning (DRL)-based power control agents on the GPU embedded software defined radios (SDRs). To this end, we propose an end-to-end framework (MR-iNet Gym) where the simulation suite and the embedded SDR development work cohesively to overcome real-world implementation hurdles. To prove feasibility, we consider the problem of distributed power control for code-division multiple access (DS-CDMA)-based LPI/D transceivers. We first build a DS-CDMA ns3 module that interacts with the OpenAI Gym environment. Next, we train the power control DRL agents in this ns3-gym simulation environment in a scenario that replicates our hardware testbed. Next, for edge (embedded on-device) deployment, the trained models are optimized for real-time operation without loss of performance. Hardware-based evaluation verifies the efficiency of DRL agents over traditional distributed constrained power control (DCPC) algorithm. More significantly, as the primary goal, this is the first work that has established the feasibility of deploying DRL to provide optimized distributed resource allocation for next-generation of GPU-embedded radios. 
\end{abstract}


\maketitle

\section{Introduction}
Wireless communication has been evolving rapidly to keep up with the requirement of the consumers both in the commercial and tactical domain. As compared to wired networks, the challenges posed by dynamic and rapidly changing channel is well known. In addition, the resources available for wireless devices are limited and do not follow the exponential growth trajectory of the number of wireless devices. These factors dictate the need for intelligent decision engines that can learn from the environment to dynamically use the scarce resources while meeting the requirements of the end customer.

The rapid growth of computational resources and availability of digital data has revived machine learning from its second winter. Machine learning has been revolutionizing several domains when applied judiciously. The wireless domain is no exception to these influences. There have been significant efforts made to effectively leverage various strengths of machine learning to solve crucial hurdles encountered in the wireless domain \cite{JagannathAdHoc2019, Luong_2019_survey, Ajagannath6G2020, RamezanpourZTA22}. For example, supervised learning is being widely deployed to perform blind signal characterization and classification
; several approaches are being leveraged to detect intrusions, anomalies, and enhance the security of wireless systems including the Internet-of-Things (IoT) 
; and reinforcement learning is being used in several applications for dynamic resource allocation. Even with the recent advances, there are several limitations and challenges that have held back the real-world deployment of machine learning based decision engines in the wireless domain. In this paper, we propose an end-to-end framework and related methodology for deployment of deep reinforcement learning (DRL) for a distributed wireless network. To accomplish this, we choose the problem of distributed power control in low probability of intercept and detection (LPI/D) networks which play a crucial role in tactical communication. Direct sequence code-division multiple access (DS-CDMA) is especially popular for implementing a robust LPI/D physical layer due to its unique advantages, such as easy frequency management (no frequency planning) in multi-user scenarios, low peak-to-average power ratio (PAPR), and less stringent synchronization requirements as necessary for orthogonal frequency division multiplexing (OFDM). We would like to emphasize the fact that this is just one example of an optimization problem that can be solved using the proposed framework which is the focus of this work.


Deploying machine learning at the edge (on embedded platform) has always been a formidable challenge in the wireless domain. To the best of our knowledge, this is the first work that puts forth an end-to-end framework to demonstrate the feasibility of designing, training, and deploying DRL-based decision engines for \textit{next generation of embedded software defined radio (SDR) hardware.} The key contributions and impact of this work are as follows,

\begin{itemize}[leftmargin=*]
\vspace{-0.1cm}
    \item We develop the first known DS-CDMA network architecture in ns3-gym to ensure rapid training of the DRL module.
    \item Next, we perform optimization of the trained model to ensure lightweight deployment on embedded SDR to minimize the inference time and ensure real-time decision making.   
    \item The results of our over-the-air experiments indicate that DRL consumes significantly less power while maintaining high reliability by taking into account higher dimensional input yet making a real-time decision.  
    \item This is the first work that establishes the feasibility of leveraging DRL for distributed decision making in the wireless network \textit{using GPU-embedded transceiver}. We believe edge deployment of machine learning solutions will be a key enabler for future networks including but not limited to 6G. 
    \item Overall, the progress and impact of this work can arguably be abstracted from the specific optimization (resource allocation) problem discussed here. 
    
\end{itemize}

\vspace{-0.25cm}
\section{Related Work}
Machine learning has been emerging as a tool suited to solve several challenging problems in the domain of wireless communication \cite{Luong_2019_survey, Ajagannath6G2020, Jagannath19MLBook, JagannathAdHoc2019}. Yet, there are several hurdles that need to be overcome to transition some of these solutions from concept to simulations to actual hardware implementation. The majority of the advances have been limited to simulations due to the challenges associated with demonstrating the effectiveness of machine learning-based approaches on actual radio hardware. It is also important to point out that in many cases (especially in the case of supervised learning) hardware is used to collect relevant data or execute the decision of a machine learning engine but the actual computation still happens on large host PCs with ample computing resources \cite{ICAMCNet, Jagannath18ICC}. Here we argue that in several cases for machine learning to truly impact the wireless domain, techniques and solutions will have to be deployed on the devices themselves using the embedded computing resources that could include GPP, GPU, or FPGA. This argument can also be seen in the light of the importance and every increasing attention edge computing has been witnessing to meet the QoS constraints (including and not limited to latency).  


In\cite{grandhi1994constrained,li2018intelligent}, distributed constrained power control (DCPC) algorithm was introduced where the power is adjusted iteratively by DCPC given the signal-to-interference-plus-noise ratio (SINR). In this approach, i.e., DCPC, a minimum required SINR threshold is considered and the optimization process continues as long as the minimum required quality of service (QoS) is achieved. Later, several other algorithms \cite{islam2007distributed, xiao2003utility,elbatt2004joint,tadrous2010power} were introduced which were the modified version of \cite{grandhi1994constrained}. In\cite{islam2007distributed}, a power and admission control algorithm with beamforming is introduced where the coexistence of the primary users (PUs) and secondary users (SUs) in a shared spectrum is guaranteed while the total transmission power is minimized. The hard SINR constraint issue in distributed power control is addressed in\cite{xiao2003utility} where a utility based power control (UBPC) is presented. In this approach\cite{xiao2003utility}, the problem is reformulated using a softened SINR constraint (utility) and a penalty for power consumption (cost) where the goal is to maximize the net utility which is equal to utility without the cost. In\cite{elbatt2004joint}, a twofold multiuser interference management approach for wireless ad hoc networks is introduced where the objectives are to increase single-hop throughput and reduce the power consumption. In this approach\cite{elbatt2004joint}, distributed power control is performed to achieve the admissible power vector which can be used by the scheduled transmitters to satisfy their single-hop requirements. A distributed power allocation algorithm is presented in\cite{tadrous2010power} where the PUs communicate under a minimum QoS requirement, whereas the SUs opportunistically utilize the primary band. Fractional programming (FP)\cite{shen2018fractional} and weighted minimum mean square error (WMMSE)\cite{sun2018learning} are the two iterative centralized power control algorithms which are model driven \cite{nasir2020deep}.
In the two aforementioned centralized power control algorithms, i.e., FP and WMMSE, the delay caused by the feedback exchange mechanism between a central controller and the users \cite{nasir2020deep} could deteriorate performance in a distributed LPI/D network. On the other hand, the existing power control algorithms require perfect CSI knowledge and are model driven. The data driven methods, e.g., DRL do not require \textit{perfect CSI knowledge}, hence, they are more promising for realistic wireless environments. To tackle the challenges of conventional power control algorithms, in this paper for the first time, we train and deploy the DRL agents on GPU-embedded SDRs. In particular, we train and deploy actor-critic (AC) networks with deep deterministic policy gradient (DDPG) for power control in a distributed network of GPU-embedded AIR-T SDRs. \textit{Additionally, all the recent works discussed above have been limited to only simulations.}
\section{Problem Formulation}
\subsection{An overview of Reinforcement Learning}
In a general Markov decision process (MDP), a single or multiple agents interact with their surrounding environment. For each action taken by the agents, a feedback reward and the new state is returned by the environment. At each given time step, $t$, the environment's essential features or state space (observation space), $s^{(t)} \in \mathcal{S}$, is observed by the agent where $\mathcal{S}$ is the set of all possible states. The agent picks an action $a^{(t)}$ from the set of all possible actions, $\mathcal{A}$, based on a policy which could be either deterministic and stochastic. The deterministic and stochastic policies are denoted by $\mu$, and $\pi$, respectively, where $a^{(t)}=\mu(s^{(t)})$ or $a^{(t)}=\pi(s^{(t)})$. After the $a^{(t)}$ is taken by the agent, the environment's state moves to a new state, namely, $s^{(t+1)}$. The transition probability from state-action pair, i.e., $<s^{(t)},a^{(t)}>$ to the new state $s^{(t+1)}$ is denoted by $P^{a^{(t)}_{s^{(t)}\longrightarrow s^{(t+1)}}} = Pr(s^{(t+1)}|s^{(t)},a^{(t)})$. The reward signal, $r^{(t+1)}$, which is returned from the environment, determines how good or bad the action is. The set of described interactions is called as an experience at $t+1$ and is denoted as $e^{(t+1)}=(s^{(t)}, a^{(t)}, s^{(t+1)}, r^{(t+1)})$.

In model free reinforcement learning, these interactions are learned without any prior knowledge about the transition probabilities and the optimal policy is achieved through maximizing the long-term accumulative discounted return of agent at time $t$, 
\begin{equation}
G^{(t)} = \sum_{i=0}^{\infty} \gamma ^{i} r^{(t + i + 1)},
\label{eq1chap4}
\vspace{-0.1cm}
\end{equation}
where $\gamma < 1$ denotes the discount factor. 
DDPG is a well known actor-critic based technique which supports continuous action spaces. In a DDPG network, a critic network is trained iteratively which learns the action value function. The optimal deterministic policy is determined by another network, namely, actor.
\subsection{Power Control Algorithm}

We assume $K$ nodes (wireless transceivers) with spreading sequence matrix $\mathbf{S}\in \mathbb{C}^{L\times K}$ forming $K/2$ transmit-receive pairs operating simultaneously in the same frequency and $L$ corresponds to the length of the spreading codes. 

To maintain distributed operation and minimize overhead, we use only local information and minimal information gathered from immediate neighbors. The state of a node $i$ at time slot $\tau$ can be expressed as
\begin{align}
    \mathcal{S}_i^{\tau} &= \{ \mathbb{D}_i^{\tau}, 
    B_i^{\tau},
    I_i^{\tau}, 
    S_i^{\tau} \}
\end{align}
where $\mathbb{D}_i^{\tau} = \{d_{ij}^{\tau} \vert j=1,2,\cdots,K \}$ is the set of distances to neighboring receivers,
$B_i^{\tau}$ is the number of packets in the buffer, $I_i^{\tau}$ is the interference caused by a transmission from node $i$ (can be estimated by node $i$), and $S_i^{\tau}$ is the interference sensed by the receiver of node $i$. The value for interference sensed at the receiver is communicated to node $i$ through an ACK. If the transmission from node $i$ is not received, then it will not receive an ACK message and the value for $S_i^{\tau}$ is set to -1. The action of node $i$ at time slot $\tau$ is given by 
 \begin{align*}
     \mathcal{A}_i^{\tau} = \{a_i \vert a_i \in \mathbb{A} \}
 \end{align*}
where the action space is the set of available power levels $\mathbb{A} =  [0.1 \text{mW}, 5\text{mW}]$. This is a hardware specific choice and can be customized when the target hardware capabilities change.  

We implement a reward function which aims to maximize SINR on successful transmission and penalize based on interference caused for unsuccessful transmission (\emph{i.e.,}\; transmission does not complete with an acknowledgement). The SINR of a transmission from node $i$ to node $j$ at a given slot $\tau$ is represented as $\Gamma_{ij}^\tau$. Let us define the interference caused by a transmission from agent $i$ to agent $j$ be $I_{i j} = \sum_{k \neq j \neq i}\frac{a_{k}}{d^{\alpha}} $ where the summation index $k$ goes over all neighboring nodes where $d$ is the distance and $\alpha$ is the path loss coefficient.

The reward for taking an action $\mathcal{A}_i$ is given below.
\begin{align}
 \label{eqreward}
    R_i = \begin{cases}
    \tilde{\Gamma_{ij}} \text{if transmission successful} \\
    - \tilde{I_{ij}}, \text{if transmission unsuccessful}
    \end{cases}
\end{align}

Where $\tilde{\Gamma_{ij}}$ and $\tilde{I_{ij}}$ are, respectively, the normalized SINR and  normalized interference caused from a transmission from agent $i$ to agent $j$. We normalize these quantities by experimentally finding the largest value likely to be observed and then using this value as a divisor. We emphasize that this is only one formulation of a resource allocation problem that can leverage the proposed framework. It can be easily seen how this could be applied to a plethora of wireless network optimization problems.  

\begin{figure*}[!ht]
\begin{minipage}[h]{0.5 \linewidth}
\centering
\centerline{\includegraphics[width=0.99\columnwidth]{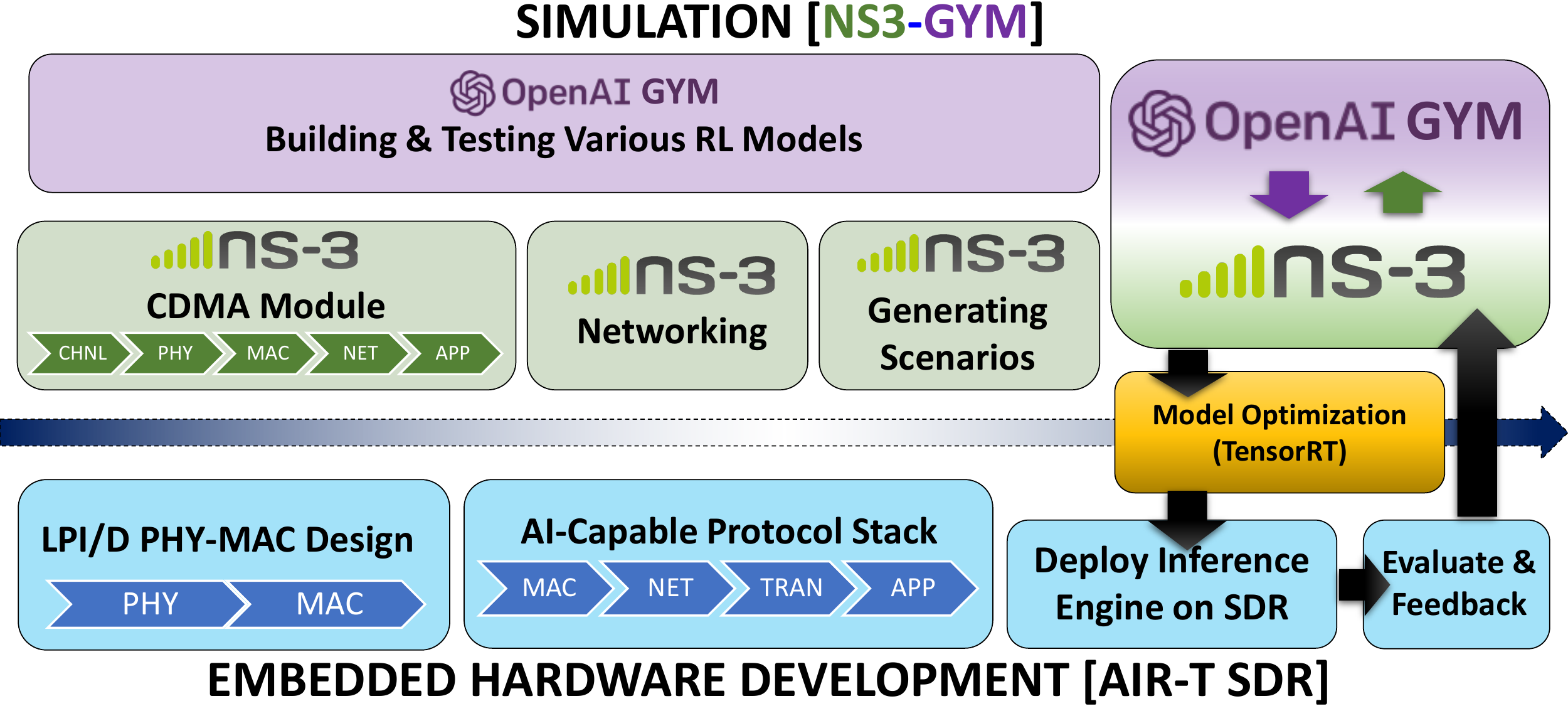}}
	\caption{MR-iNet Gym Framework for Intelligent Networks}
	\label{fig:toolchain}
\end{minipage}
\hspace{0.2 cm}
\begin{minipage}[h]{0.4 \linewidth}
\centerline{\includegraphics[width=0.99\columnwidth]{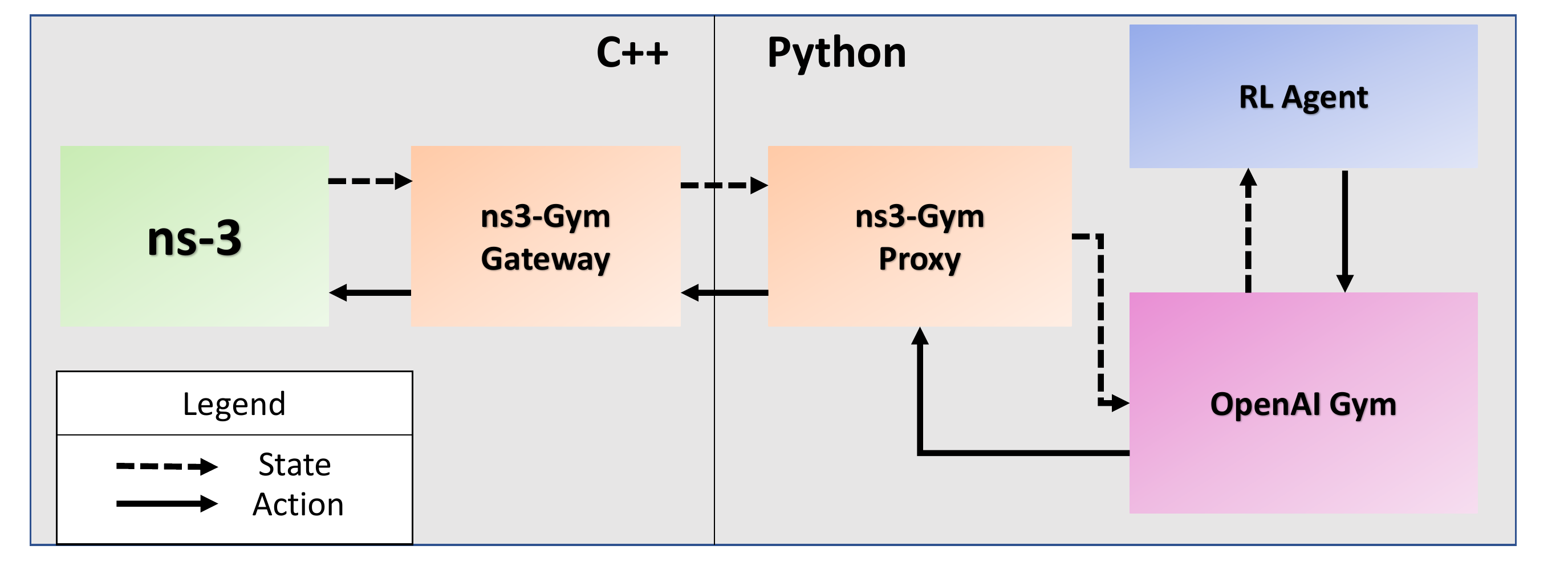}}
	\caption{ns3-gym information flow: Dotted lines indicate the transfer of state information. Solid lines indicate the communication of the action selected by the RL agent.}
	\label{fig:ns3-gym}
\end{minipage}
\hspace{0.2 cm}
\vspace{-0.3 cm}
\end{figure*}
\vspace{-.1 cm}

\section{MR-iNet Gym Framework}\label{sec3}
In this work, we present the Marconi-Rosenblatt framework for intelligent networks (MR-iNet Gym) shown in Fig. \ref{fig:toolchain}. The MR-iNet Gym facilitates the designing, training, and deployment of the DRL decision engine on embedded SDR. To accomplish the objective of deploying DRL on embedded transceiver hardware, we identify three main components, (i) A high fidelity simulator that facilitates design and setup of network scenarios while allowing the transceivers (agents) to run DRL decision engines with ease, (ii) methodology to optimize (from computation standpoint) the trained model for embedded deployment on wireless transceivers, and (iii) a flexible embedded software defined radio hardware that can house the AI-enabled communication protocol stack. 
\vspace{-.1 cm}

\subsection{Simulation Suite for AI-enabled Networks}
For our simulation environment, to ensure rapid development, we leverage several open source software packages for C++ and Python. We begin by building the first known custom DS-CDMA module for ns-3 to simulate a distributed LPI/D wireless network. Next, we use OpenAI Gym \cite{Gym} to handle the reinforcement learning training loop. This enables us to rapidly implement several DRL algorithms to compare and fine-tune the best-performing model. In this work, we focus on DDPG with continuous action space as it maps naturally onto the power control problem. Finally, we use ns3-gym \cite{ns3gym} to serve as a gateway between the simulation environment provided by ns-3 with reinforcement learning structure provided by Gym. Inside of Gym training loop our reinforcement learning agent then receives state information, uses this information to select an action, and then passes the selected action back to the environment. The information flow between ns-3 and Gym that ns3-gym facilitates is shown in Fig. \ref{fig:ns3-gym}. 
\subsection{Target Embedded Software Defined Radio}

In the conventional embedded SDRs,  General Purpose Processing (GPPs) or Field  Programmable  Gate  Arrays  (FPGAs) are commonly used to provide the processing capabilities. Neither of these is ideal for deploying DRL decision engines and often leads to undesired latency. Therefore, deploying the deep learning (DL)-based solutions on the SDRs will even further reduce their speed which makes the adoption of the conventional SDRs in the real-world applications impractical in the fifth generation (5G) and beyond communication systems. Therefore, it is required to utilize the parallel processing computational power of the graphical processing units (GPUs) in the embedded SDRs for: 1) offloading the processing of GPP/CPU onto GPU to accelerate the signal processing; 2) making it practical to deploy the DL-based solutions while not affecting the speed of the signal processing algorithms through the parallel processing capabilities of the embedded GPU. Due to the aforementioned reasons, in this work, we implement and deploy our signal processing operations and AC power control engines on the \textit{Deepwave Digital's Artificial Intelligence Radio Transceiver (AIR-T) which is the world's first SDR equipped with embedded GPUs}. The AIR-Ts (shown in Fig. \ref{Fig:AIR-T}) are designed and developed to operate in the DL-based radio frequency (RF) applications and they are equipped with embedded NVIDIA GPU, an FPGA, and dual embedded CPUs. 

\vspace{-.1 cm}
\subsection{Bridging the Reality Gap}

Another key roadblock to overcome is the reality gap between simulations and hardware. Even with reliable simulations, it is extremely difficult to capture the unpredictable behaviors of the wireless channels and its instantaneous effects on the physical layer of modern communication systems. While certain abstractions at the physical layer are appropriate and highly beneficial in reducing the complexity of the simulator, it is important those abstractions happen in an informed manner trying to represent hardware-specific nuances and model the interaction with the expected channel conditions as closely as possible. In other words, it is important to feedback the finding regarding generalized hardware-specific operation characteristics from hardware evaluation back into the simulation suite to minimize the degradation that can be caused by these reality gaps. This interaction is represented by the interaction between simulation and hardware deployment in Fig. \ref{fig:toolchain}.

\section{Simulation and Testbed Experiments}
\begin{figure}[ht]
\begin{center}
\centerline{\includegraphics[width=\columnwidth, height=0.25\columnwidth]{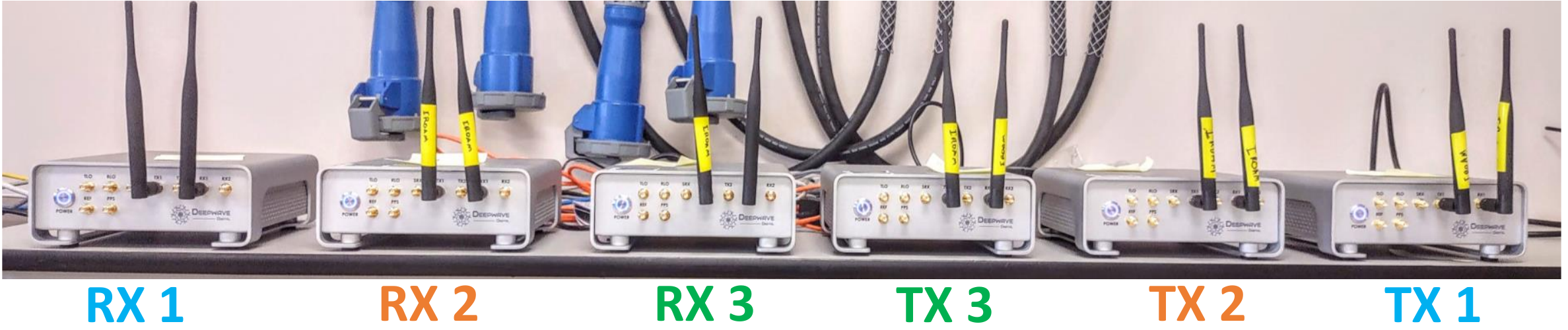}}
\caption{6-node GPU-embedded SDR testbed.}
\label{Fig:testbed}
\end{center}
\vspace{-0.33in}
\end{figure}

Our testbed shown in Fig. \ref{Fig:testbed} consists of six AIR-T SDRs which in total establish three CDMA-based communication pairs. Due to the space restriction, we do not elaborate on the physical layer implementation but the solution can be extended to any CDMA based tactical or commercial network. We implement \textit{AI-enabling protocol stack} on SDR to ensure communication and interfaces between layers so that information can flow from different layers to the DRL engine. The DRL agents are deployed on each transmitter and the power control decisions are made by the agents after they receive feedback from their corresponding receiver.

\subsection{Training in NS3-GYM}

We train our AC power control agents with DDPG algorithm in ns3-gym environment. 
The AC networks constitute an actor network and a critic network. After trying different configurations, the final actor network is a fully connected DNN with one hidden layer where there are 10 neurons in the hidden layer. We use rectified linear unit (ReLU) used as the activation function in all the layers. The critic network is a fully connected DNN with four hidden layers where there are 16, 32, 32, and 256 neurons in each hidden layer, respectively. Similarly, ReLU function is adopted as the activation function in all layers. The training is performed using the reward function given in Eq~\ref{eqreward} and the reward plot of the AC agents trained by DDPG algorithm in ns3-gym environment is plotted in Fig.~\ref{iccn6}. As it can be seen in Fig.\ref{iccn6}, the reward plot of the AC agents respectively converging within 2000 steps.  However, as it can be seen, there is a high variance between different runs for the agents while the environment has not changed. We will address this issue in section~\ref{MADDPG} by introducing an approach to improve the stability of DDPG actors. 
\vspace{-0.5cm}
\subsection{Stabilizing DDPG With Model Aggregation}\label{MADDPG}

\begin{figure*}[!h]
\begin{minipage}[h]{0.45 \linewidth}
\centering
\centerline{\includegraphics[width=\columnwidth]{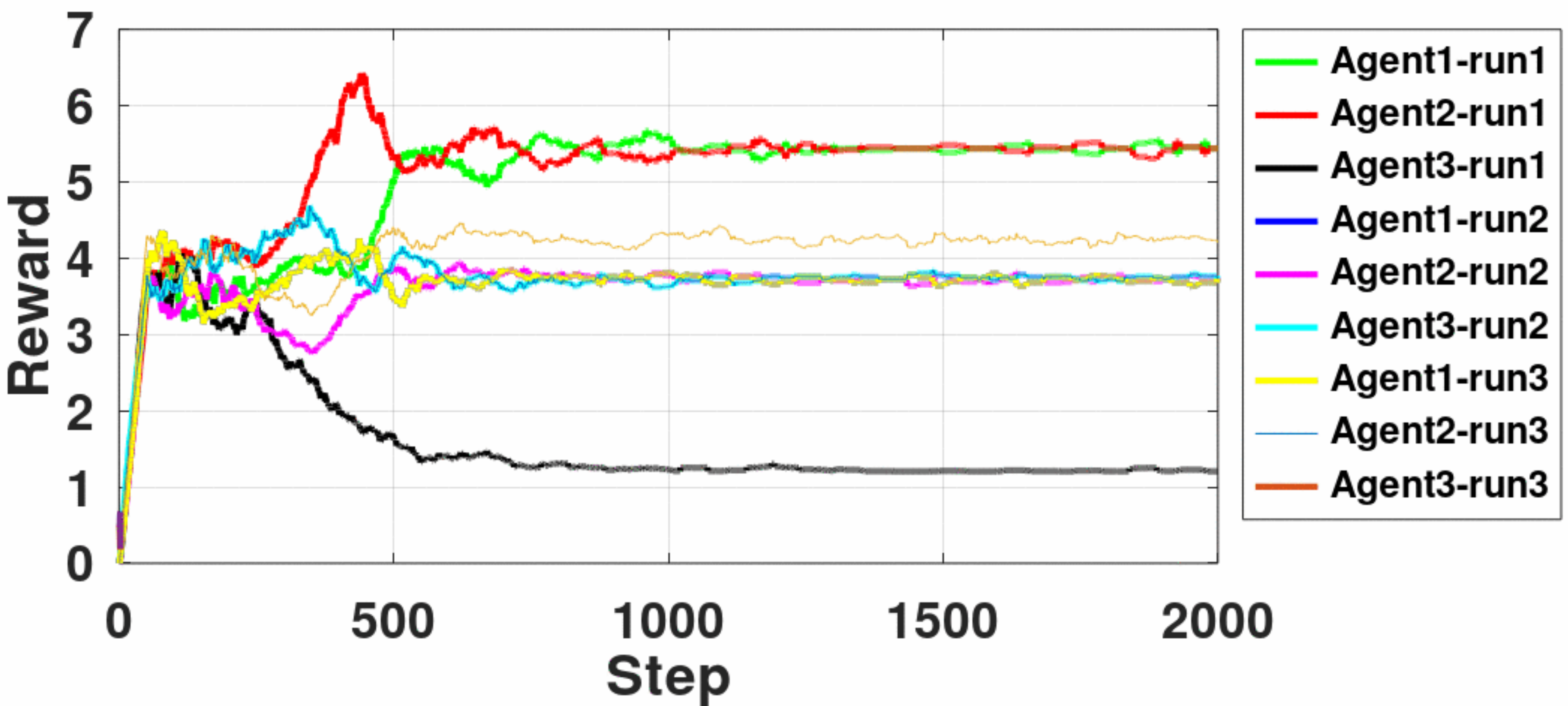}}
	\caption{Agents' rewards w/o model aggregation}
	\label{iccn6}
\end{minipage}
\hspace{0.2 cm}
\begin{minipage}[h]{0.47 \linewidth}
\centerline{\includegraphics[width=\columnwidth]{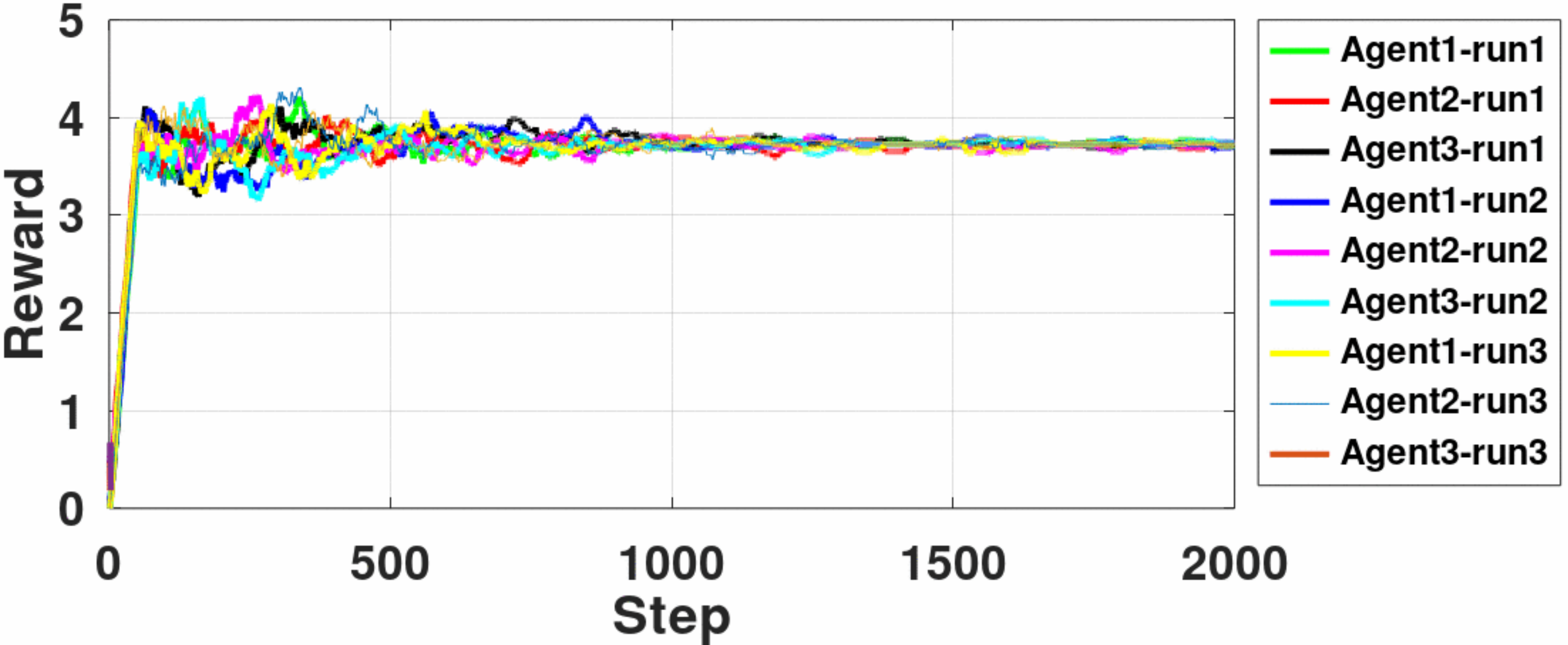}}
	\caption{Agents' rewards with model aggregation.}
	\label{iccn7}
\end{minipage}
\hspace{0.2 cm}
\vspace{-.3cm}
\end{figure*}

In a DDPG network, the policy is determined by a non-linear neural network which is very sensitive to its parameters, namely, $\theta$. This implies that a small variation in $\theta$ could result in a very large difference in the policy that is represented by the actor network. In Figure~\ref{iccn6}, the reward plots of three agents on three different runs on the same environment have been demonstrated. As it can be seen, the variance of the rewards among different runs for the same agent is very high. In other words, although our environment has not changed, the DDPG agents are converging to different points amongst different runs. This behavior implies that the DDPG agents are not demonstrating a good stability. This is mainly caused due to the non-convex optimization problem, where a small change in the gradient steps can lead to a different policy which could be arbitrarily a bad solution\cite{tessler2019stabilizing}. In general, policy-based DRL methods such as DDPG perform poorly when compared against tabular methods in terms of stability.

Therefore, in this paper, we introduce an approach which improves the stability of policy-based DRL methods, in particular DDPG network. We apply our approach on DDPG network and verify its effectiveness through extensive simulations. However, without loss of generality, this approach can be applied on any policy-based DRL method. The approach that we propose to stabilize the DDPG network is based on model aggregation. Combining different models together is a well know approach in machine learning to improve the robustness and decrease the variance of the models' outputs. In this way, the sensitivity of the models to initial parameters and noise is decreased. Algorithm 1 is the pseudo code representing our approach for model aggregation and stabilizing of DDPG agents.  

\begin{figure}[ht]
\begin{center}
\centerline{\includegraphics[width=\columnwidth]{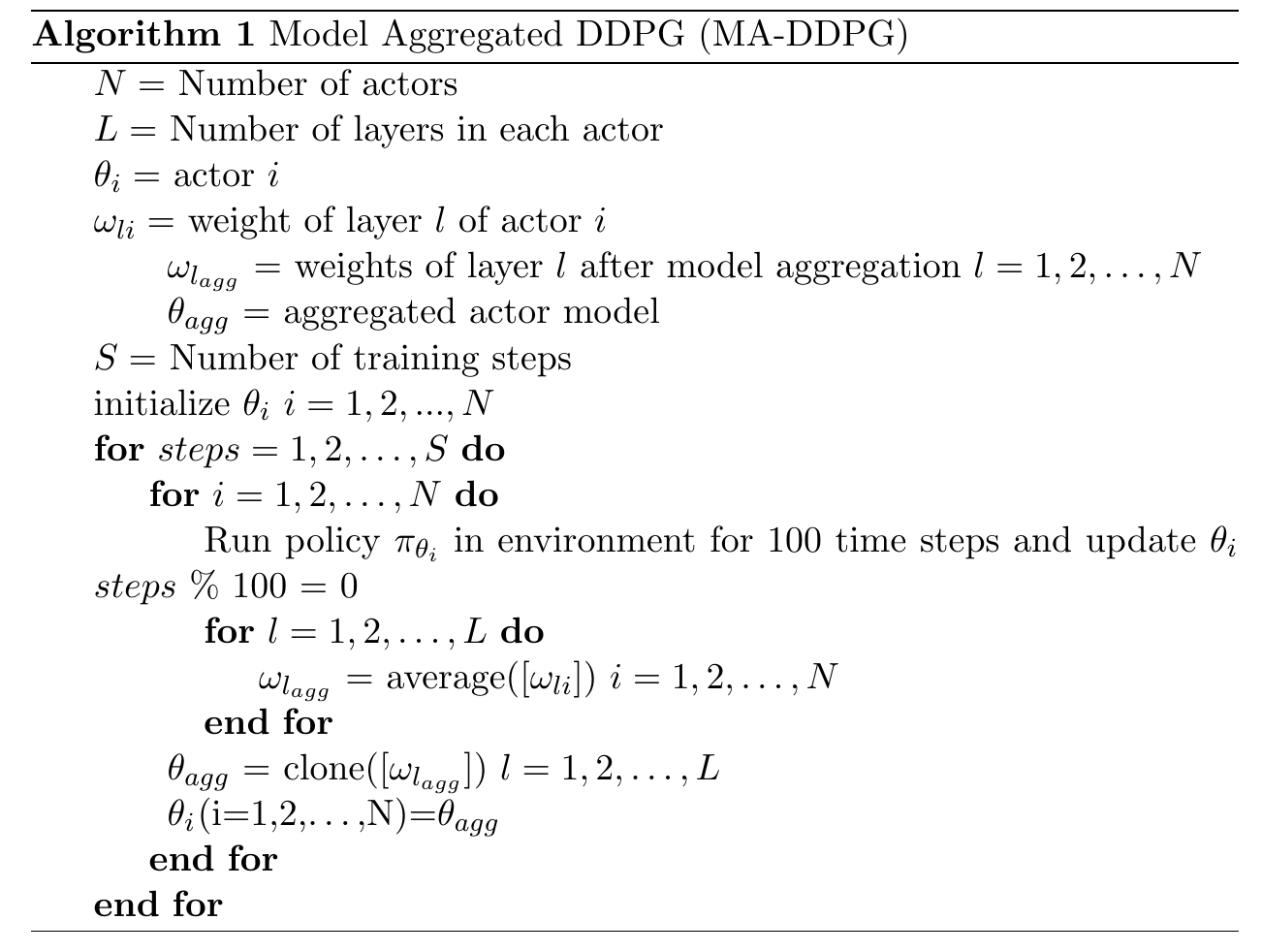}}
\end{center}
\vspace{-.98 cm}
\end{figure}
We perform the model aggregation of the DDPG actors every 100 steps. As it can be seen in Fig ~\ref{iccn7}, after performing the model aggregation, the actors show much less variance during the training among several runs compared to the case where no model aggregation has been performed, i.e., Fig~\ref{iccn6}. 
\subsection{Optimizing the DRL Model for Hardware}\label{sec:hardware_optimization}

\begin{figure}[ht]
\begin{center}
\centerline{\includegraphics[width=\columnwidth]{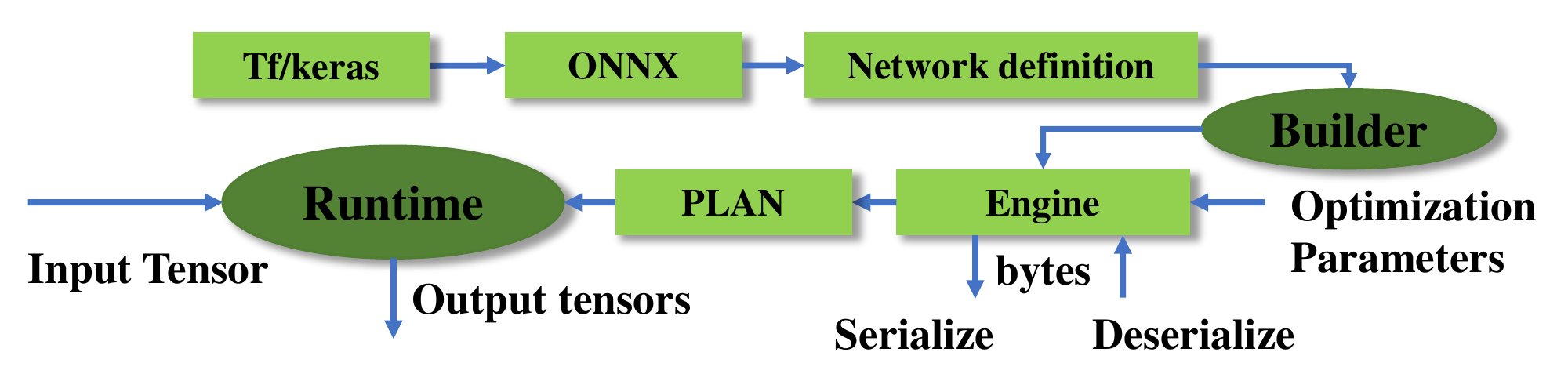}}
\caption{Flowchart of TensorRT model optimization.}
\label{icc1}
\end{center}
\vspace{-0.5 cm}
\end{figure}

To ensure real-time operation, it is essential to accelerate the decision making process of the DRL-based inference engines on the SDRs. Our preliminary results indicate that adopting the DNN framework directly on the SDRs, can cause high latency in the delivery of the packets, hence it is not an optimal approach for a delay-sensitive application. In our SDR hardware setup, the Nvidia Jetson TX2 module is the accelerating hardware module and the TensorRT is the software framework for accelerating the DNN inference engines. TensorRT is comprised of an inference optimizer and a runtime which together deliver low-latency and high-throughput for resource allocation on SDRs. In Fig.~\ref{icc1}, the flowchart of converting the DNN models to the optimized TensorRT format is depicted.

\begin{figure}[h!]

\begin{minipage}[h]{0.33 \linewidth}
\vspace{-0.5cm}
\centerline{\includegraphics[width=0.9\columnwidth]{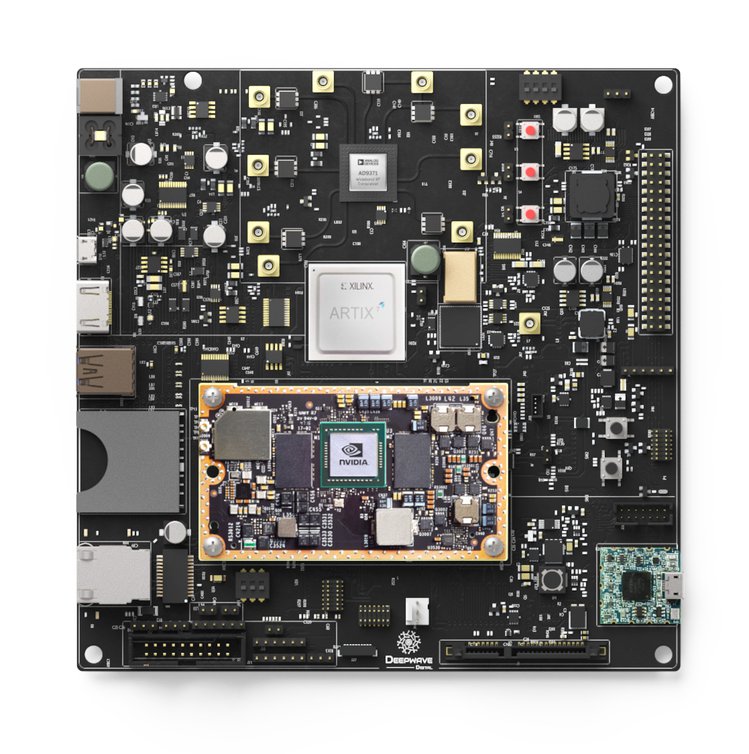}}
\vspace{-0.3cm}
\caption{AIR-T SDR} 
\label{Fig:AIR-T}
\end{minipage}
\hspace{0.5 cm}
\begin{minipage}[h]{0.55 \linewidth}
	 \small
	\centering 
	\begin{tabular}{c r} 
		\hline\hline 
		Format&\multicolumn{1}{c}{Inference time} \\ [1ex]
		\hline 
	Original & 2266 $\mu$s \\[1ex] 
		ONNX & 14.6 $\mu$s \\[1ex]
        PLAN & 0.23 $\mu$s\\[1ex] 

		\hline 
	\end{tabular}
	\captionof{table}{Inference time of a DNN model in different formats}
	\label{tab:inference}
	
\end{minipage}
\vspace{-0.9cm}
\end{figure}

Firstly, the Tensorflow based DNN model of the DRL network which is trained in the ns3-gym environment is saved. Consequently, this model is converted to the ONNX format. Then, the inference engine is built by TensorRT which can be serialized to PLAN file. Our SDR-based resource allocation method utilizes TensorRT to control the GPU-based DNN inference engine.  In this paper, for the first time, the performance of a SDR-based power allocation algorithm which relies on TensorRT and CUDA to optimize the DNN inference engine is presented. In Table~\ref{tab:inference} a comparison between the original model(.h5 format), ONNX format and PLAN (optimized by TensorRT) is demonstrated with respect to the average inference time of 30000 runs. The inference time of the original model, and ONNX model are measured on GPP. However, the inference time of the PLAN file (optimized by TensorRT) is measured on the GPU (cannot run on GPP).  As the results show, the total inference time of the optimized model is significantly lower than the original, and ONNX models, respectively, while the precision of model has not been compromised. 

\subsection{DRL Over-the-air Results}

\begin{figure*}[h!]
\begin{minipage}[h]{0.37 \linewidth}
\centering
\centerline{\includegraphics[width=0.99\columnwidth]{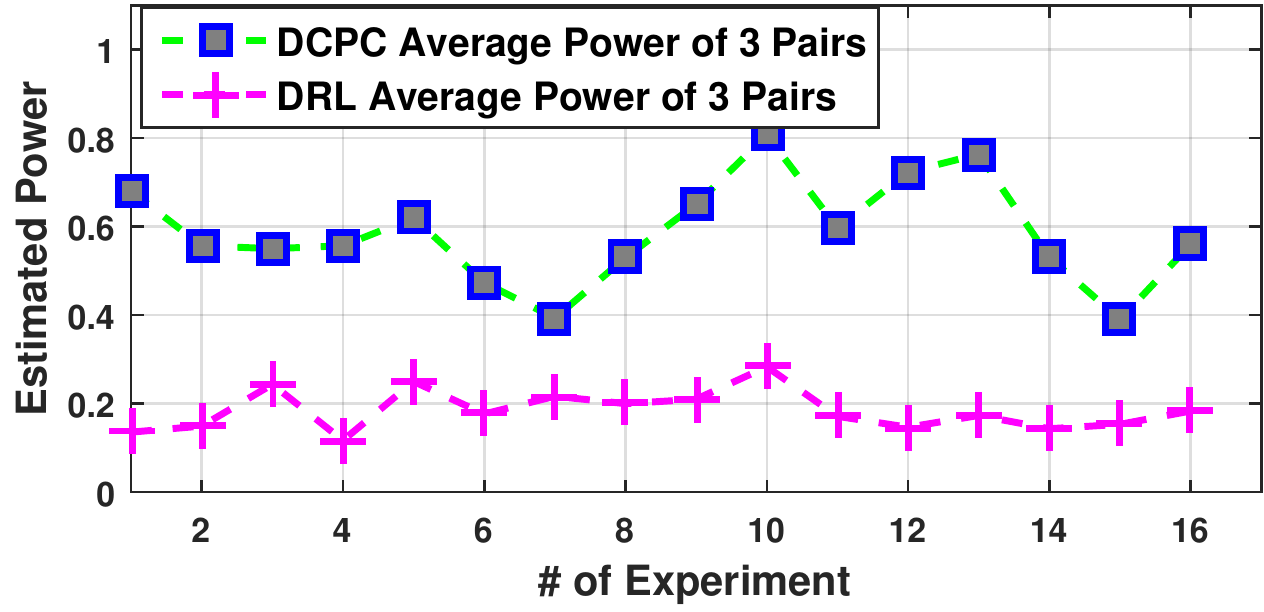}}
\caption{Estimated power for DRL and DCPC.}
\label{icc6}
\end{minipage}
\hspace{0.2 cm}
\begin{minipage}[h]{0.37 \linewidth}
\centerline{\includegraphics[width=0.99\columnwidth]{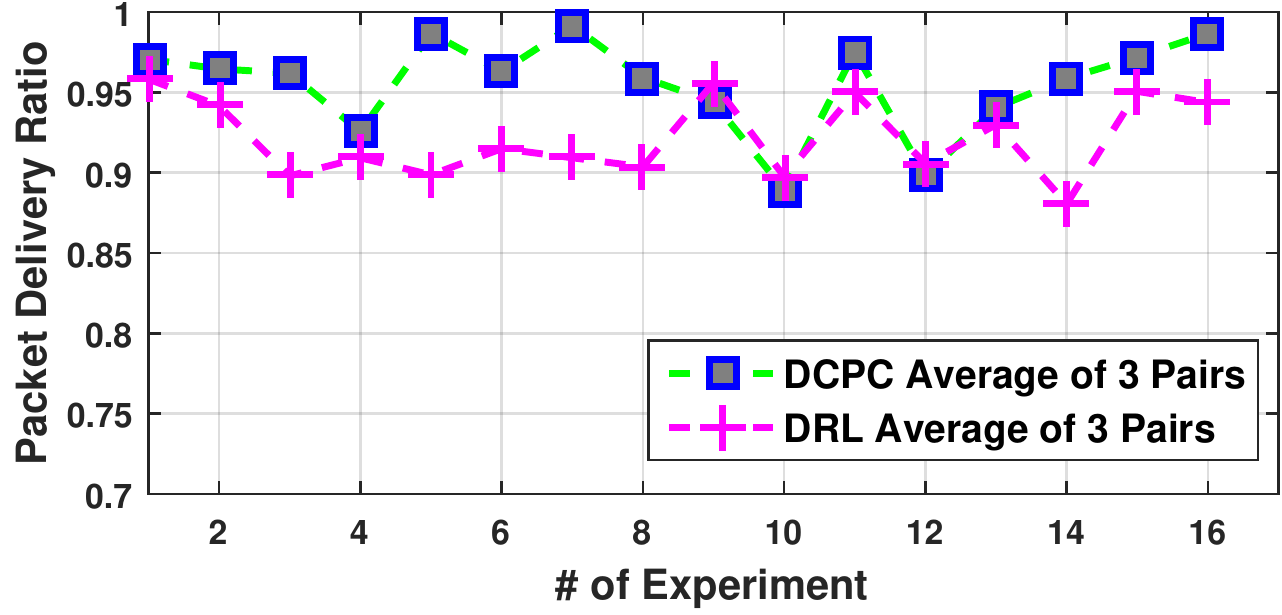}}
\caption{The average PDR for DRL and DCPC.}
\label{icc7}
\end{minipage}
\hspace{0.2 cm}
\begin{minipage}[h]{0.22 \linewidth}
\centering
\small
\begin{tabular}{c c r} 
		\hline\hline 
		Format& Power&\multicolumn{1}{c}{PDR} \\ [1ex]
		\hline 
	DRL & 0.19mW & .922\\[1ex] 
		DCPC & 0.59mW & .955 \\

		\hline 
	\end{tabular}	\label{tab:comparison}
	\vspace{0.3cm}
	\captionof{table}{Performance Comparison of DRL agents and DCPC Algorithm} 
	 \small
\end{minipage}
\vspace{-0.2 cm}
\end{figure*}

To deploy the models on the AIR-T SDRs, the PLAN files are required to be linked to the DS-CDMA physical layer to perform inference. Our DS-CDMA physical layer is developed purely in C++, however, the PLAN inference engines are CUDA and GPU friendly. To manage the complexity in a large program like ours, it is essential to break down the large program into smaller components. In fact, the CUDA-based PLAN inference engines and C++ codes are compiled separately and are linked together using the linking tools that have been made available since CUDA 5.0. The experiments were conducted in the 2.45 GHz ISM band. The power of the first packet is set by the user (can be set to the max value to ensure reliability). Upon the successful delivery of the first packet at the receiver, an ACK (acknowledgment) message is transmitted from the receiver to the transmitter which contains two key elements (i) SINR, and (ii) the interference sensed at the receiver. The information required for the DRL which included SINR, interference, buffer length, and the distance of each transmitter from its intended receiver and other receivers as well, are fed to the DRL agents to choose optimal action (optimal power between 0 to 5 mW) for the transmission of the next packet. 
For each experiment 1000 packets were transmitted and in total 16 experiments were performed where each experiment corresponded to a different combination of the transmitter and receiver gains to represent different operation constraints. To evaluate the performance of the DRL agents, we compare their performance against DCPC\cite{li2018intelligent,grandhi1994constrained}, in terms of PDR and total power consumption.           

Fig.~\ref{icc6} depicts the estimated average power of 1000 packet transmission for different experiments. As it was mentioned, each experiment corresponds to a different combination of the transmitter and receiver gains. Fig.~\ref{icc7} shows the average PDR of DCPC and DRL. The power consumption of DCPC is significantly higher than of DRL without a significant gain in PDR. In Table II, the average performance metrics of DCPC and DRL are presented. The results indicate that the power consumption of DCPC is $\approx 3$ times higher.

\vspace{-0.2cm}
\section{Conclusion and Future Work}
Machine learning has been rapidly emerging as a tool to accelerate the strides towards realizing the next-generation of radios. While there have been several advancements in this domain, the majority of them have been limited to simulations or relied on host-PC with ample computational resources. Due to the inherent need for mobility, energy-efficiency, and other strict Quality-of-Service requirements, it is inevitable that machine learning models will have to live on embedded devices (at the edge) to ensure AI-enabled radios can make real-world impacts. In this work, for the first time in literature, we have used GPU-embedded SDR to demonstrate real-time (<$\mu$s) distributed decision making for a resource allocation problem in distributed wireless networks. To accomplish this, we proposed the MR-iNet Gym framework that orchestrates a cohesive development process. This leverages ns3-gym simulation for training, optimizing the trained model for hardware deployment, and achieving superior performance on GPU-embedded SDR hardware in our over-the-air experiments.  

In the future, we hope to publish an extended version discussing the physical layer and AI-enabling protocol stack implementation in detail. In 2022, we also hope to release the ns3-gym software since it is the first one developed for the CDMA network. We are actively utilizing MR-iNet Gym to address a wide variety of network optimization problems which we hope to discuss in our future work. Finally, we hope the contribution of this paper has a broader impact that motivates the community to adopt the design ethos of MR-iNet Gym to solve open research challenges in the wireless domain.
\section*{Acknowledgment and Disclaimer}

This material is based upon work supported by the US Army Contract No. W15P7T-20-C-0006. Any opinions, findings, and conclusions or recommendation expressed in this material are those of the author(s) and do not necessarily reflect the views of the US Army.
\bibliographystyle{ACM-Reference-Format}
\bibliography{sample-base}

\end{document}